\documentclass[11pt,a4paper]{article}
\usepackage[table]{xcolor}
\usepackage{setspace}
\usepackage[hyperref]{acl2021}
\usepackage{times}
\usepackage{latexsym}
\usepackage{amsmath}
\usepackage{amssymb}
\usepackage{amsthm}
\usepackage{bbold}
\usepackage{inconsolata}

\usepackage[capitalize]{cleveref}
\usepackage{comment}
\usepackage{bm}

\newcommand\expect{\mathbb{E}}
\newcommand{\V}{\mathcal{V}}

\theoremstyle{definition}
\newtheorem{defn}{Definition}

\usepackage{microtype}

\aclfinalcopy
\def\aclpaperid{2769}

\usepackage[disable]{todonotes}

\newcommand{\viztwo}[5]{

\vspace{1mm}
\noindent
\hspace{-3.5mm}
\arrayrulecolor[RGB]{#3,#4,#5}
\setlength\arrayrulewidth{1.2pt}
\renewcommand\arraystretch{1.1}
{
\footnotesize
\begin{tabular}{|p{\linewidth - 5mm}|}
\hline
\cellcolor[RGB]{#3,#4,#5} \bf #1 \\
\vspace{-1.3mm}
\raggedright\setstretch{.75}\it #2 
\vspace{1mm}
\tabularnewline
\hline
\end{tabular}
}
}

\usepackage{subfig}

\title{What Context Features Can Transformer Language Models Use?}

\author{Joe O'Connor ~~~~ Jacob Andreas \\
  Massachusetts Institute of Technology \\
  \texttt{\{joeoc,jda\}@mit.edu} 
}

\date{}

\begin{document}
\maketitle
\begin{abstract}

Transformer-based language models benefit from conditioning on contexts of hundreds to thousands of previous tokens. What aspects of these contexts contribute to accurate model prediction? We describe a series of experiments that measure \emph{usable information} by selectively ablating lexical and structural information in transformer language models trained on English Wikipedia. In both mid- and long-range contexts, we find that several extremely destructive context manipulations---including shuffling word order within sentences and deleting all words other than nouns---remove less than 15\% of the usable information. Our results suggest that long contexts, but not their detailed syntactic and propositional content, are important for the low perplexity of current transformer language models.\footnote{Code for all experiments in this paper is available at \url{https://github.com/lingo-mit/context-ablations}.}

\end{abstract}

\section{Introduction}
\label{sec:introduction}

Recent years have seen a significant improvement in the predictive accuracy of neural language models (LMs), owing to a combination of improvements in model architecture (especially transformers; \citealt{transformer}) and training infrastructure \citep{huggingface}. The most striking change, relative to both recurrent neural LMs \citep{mikolov2010recurrent} and count-based models \citep{kneser1995improved}, is the length of the context that these models can effectively condition on. While count-based LMs in production speech recognition and machine translation systems typically used 10--20 tokens at a maximum \citep[e.g.,][]{brown2011cmu}, and recurrent LMs have an effective context size of 200 \citep{sharp-fuzzy}, the predictive accuracy of transformer LMs appears to improve when conditioning on as many as a thousand previous tokens \citep{longformer}. A significant amount of recent work has focused on making use of even longer contexts computationally feasible \cite{compressive,linformer,sparse,transformer-xl,reformer}.

But despite empirical evidence that long contexts are helpful, little is understood about \emph{why}. If the future of language modeling will include a focus on contexts of increasing size, it is important to first understand \emph{what contextual information contributes to accurate prediction} in current models. This paper offers an answer to that question via the \textbf{$\bm{\V}$-information} framework of \citet{usable-information}. $\V$-information, discussed more in \cref{sec:approach}, provides a formal framework for reasoning about how much \emph{usable information} a computationally constrained predictor (like a neural LM) can extract from an input. Our experiments measure the amount of usable information that is added when increasing LM context size, then attempt to pinpoint the \emph{source} of this information by ablating features of the added context (via controlled shuffling and word deletion) and measuring the resulting loss of model predictive power. While this framework is general, we focus on transformer LMs.

Our work is closely related to an earlier study by \citet{sharp-fuzzy}, which measured changes in a pre-trained LSTM LM when context words were permuted and deleted at evaluation time. But neural language models are known to be highly sensitive to distributional shifts---and in particular might be unable to use information from long-range context but still be adversely affected when the structure of that context changes at evaluation time. Directly measuring usable information makes it possible to clearly distinguish accuracy decreases that result from \emph{loss of information} and decreases that result from \emph{out-of-distribution inputs}.

Our experiments reveal a number of surprising facts about the use of long- and mid-range context in transformers. While increasing context length from 256 to 768 tokens is beneficial (decreasing perplexity by roughly 4\%), many destructive transformations of this context (including transformations that cause large changes in the paradigm of \citealt{sharp-fuzzy}) remove essentially no usable information. Our results suggest that for current models, the primary carriers of information in long-range context are content words and local co-occurrence statistics: deleting function words and shuffling within local windows both have very little effect on models' predictive power. Context matters, but not all features of context matter equally; as discussed in \cref{sec:discussion}, these results motivate future language modeling research focused on alternative context representations rather than simply more tokens.

\section{Approach}
\label{sec:approach}

A \textbf{language model} (LM) places a probability distribution $p(x)$ over discrete token sequences $x$. Most learned LMs do so by decomposing $p(x)$ according to the chain rule and modeling the conditional distribution over a single \textbf{target token} given a (fixed- or variable-length) \textbf{context} of previous tokens:
\begin{equation}
    \label{eq:autoregressive}
    p(x) = \prod_{i} p(x_i \mid x_0, x_1, \ldots, x_{i-1}) ~ .
\end{equation}
In \textbf{transformer language models}, this conditional distribution is modeled via a sequence of alternating neural feed-forward layers and self-attention layers; see \citet{transformer} for more details.

While input sequences $x$ can in principle be made arbitrarily long, there are both theoretical and practical limits to transformers' ability to make effective use of it \citep{hahn-2020-theoretical,wang2019superglue}. Here, we wish to understand when (and why) increasing the size of the context improves model predictions.

\paragraph{Usable information}

Consider a hypothetical LM context consisting of the tokens \emph{The user's password is\ldots}. This context suggests that subsequent tokens will be a password: (hopefully!) a high-entropy sequence. Now suppose this context is extended to include earlier tokens, becoming \emph{The user's hashed password is ave\$@To9!. The user's password is\ldots}. Information-theoretically, this context is extremely informative: only a small number of passwords will hash to the given string, and a predictor capable of testing all passwords would be able to identify the candidates and significantly reduce its uncertainty about future tokens. But in practice, this extra context is useless: no known efficient predictor can learn anything about the password from its hash code, and the extra context has not made the language modeling problem any easier. This is an extreme case, but a similar intuition applies to more conventional questions about language models. A newspaper article whose first sentence begins \emph{A dog bit a man} is likely to end very differently from one that begins \emph{A man bit a dog}. Can LMs reason effectively about this distinction, or is it (like a hashed password) computationally inaccessible to current models?

A framework for answering questions of this kind was introduced by \citet{usable-information}:

\begin{defn}
The usable predictive information (formally, predictive \textbf{$\bm{\V}$-information}) from a random variable $X$ to a random variable $Y$ as:
\begin{align}
    \label{eq:predictive-info}
    I_\V(X \to Y) = &\big[\inf_{p_1 \in \V} -\expect \log p_1(Y)\big] \nonumber \\
    &- \big[\inf_{p_2 \in \V} -\expect \log p_2 (Y \mid X)\big]
\end{align}
for a class $\V$ of distributions $p$. 
\end{defn}
Intuitively, this definition measures how much extra information about $Y$ can be extracted from $X$ by any predictor in $\V$. In language modeling, we will take $Y$ to be the target word, $X$ its context, and $\V$ a class of parametric models. While this definition generalizes Shannon mutual information \citep{shannon1948mathematical} and has deep connections to other information-theoretic quantities (see \citealt{usable-information} for details) it ultimately corresponds to a simple and common-sense evaluation: if we want to know how much the extra context $X$ helps a language model, we should train a model $p_1$ without access to $X$, train a model $p_2$ with access to $X$, and compare the accuracy of their predictions.

\paragraph{Measuring what is used} But the original question raised by the introduction was not just \emph{how much} information is contributed by context. It is already well-established that conditioning on long contexts is helpful, with existing experiments on long-range transformers effectively implementing the measurement in \cref{eq:predictive-info}. Instead, we want to know \emph{what} information in this context is actually used by models.

As a prototypical example, let us hypothesize that more than five tokens away from the target, models are \emph{only} able to extract usable information from nouns. (In our experiments in \cref{sec:experiments}, this ``long-range context'' will be considerably longer than 5 words.) For example, given the sentence:
\begin{quote}
    \it\small Pierre Vinken, 61 years old, will join the board as a nonexecutive director Nov. 29.
\end{quote}
we hypothesize that the LM distributions:
\begin{align}
    & p_1(\textit{\small director} \mid \textit{\small Pierre Vinken, 61 years old, will} \nonumber \\[-.5em]
    & \hspace{4.9em} \textit{\small join the board as a nonexecutive}) \\[.5em]
    &\approx p_2(\textit{\small director} \mid \underbrace{\textit{\small Pierre Vinken years}}_\text{noun-only context}, \nonumber \\
    & \hspace{6em} \underbrace{\textit{\small the board as a nonexecutive}}_\text{ordinary context}) ~ ,
\end{align}
and more generally that
\begin{align}
    &I_\V(X_{0:n} \to X_n) \nonumber \\
    \label{eq:relative-info}
    &\quad \approx I_\V([\texttt{nouns}(X_{0:n-5}), X_{n-5:n}] \to X_n)
\end{align}
where $X_{i:j}$ is the sequence of tokens $[X_i, X_{i+1}, \ldots, X_{j-1}]$, $\V$ is a class of LMs, and \texttt{nouns} is a \textbf{context ablation} that extracts only the nouns from a given string. That is, we hypothesize that the amount of usable information contributed by the full context $X_{0:n}$ is the same as the amount contributed by the ablated context $[\texttt{nouns}(X_{0:n-5}), X_{n-5:n}]$, so ablation removes no information.

The experiments in this paper generalize this experimental framework to other context ablations and hypotheses. Let $f$ be an ablation and $k$ an integer offset, and denote an
\textbf{ablated context}:
\begin{equation}
    f_k(X) = [f(X_{0:n-k}), X_{n-k:n}]
\end{equation}
and 
an \textbf{ablated negative log-likelihood}:
\begin{equation}
    \label{eq:ablated-likelihood}
    \mathcal{L}(\theta, f, k) = -\expect \log p_\theta(X_n \mid f_k(X_{0:n}))
\end{equation}
Then, we can measure the effect of each ablation $f$ on usable information via the following quantity:

\begin{defn}
The \textbf{ablated information} due to an ablation $f$ at an offset $k$ is:
\begin{align}
    \mathcal{A}(f, k)  &= \frac
      {\scriptstyle I_\V(X_{0:n} \to X_n) - I_\V(f_k(X_{0:n}) \to X_n)}
      {\scriptstyle I_\V(X_{0:n} \to X_n) - I_\V(X_{n-k:n} \to X_n)} \\[0.5em]
    &= \frac
      {\scriptstyle \inf_{\theta} \mathcal{L}(\theta, f, k) - \inf_{\theta'} \mathcal{L}(\theta', n)}
      {\scriptstyle \inf_{\theta''} \mathcal{L}(\theta'', n-k) - \inf_{\theta'} \mathcal{L}(\theta', n)} ~ ,
      \label{eq:ablated-info}
\end{align}
where $\mathcal{L}(\theta, i)$ is the (unablated) negative log-likelihood $-\expect \log p_\theta (X_n \mid X_{n-i:n})$.
\end{defn}

Intuitively, $\mathcal{A}(f, k)$ measures how much of the usable information \emph{added by an extra $k$ tokens} (the denominator) is \emph{removed by applying the ablation $f$ to those $k$ tokens} (the numerator). If it is close to 0, almost no information is removed; if it is close to 1, almost all information is removed.

\paragraph{Evaluation in practice}

\cref{eq:ablated-info} provides a general framework for answering our core question in this paper: for a diverse set of context ablations and offsets, we will measure how much information is lost when a given ablation is applied at a given offset. A few modifications are required to turn this equation into a practical evaluation scheme:

\textit{Held-out evaluation}: \cref{eq:ablated-likelihood} involves an expectation over the sequence distribution $p(X)$. In practice, LMs must be trained on finite corpora, creating a risk of overfitting \citep{zhang2016understanding}. To address this issue, we approximate the infimum in \cref{eq:ablated-likelihood} by fitting $\theta_1$ on a training set, and computing ablated information on a held-out validation set. All reported results are an average of held-out likelihoods from two random initializations.

\begin{figure}
    \centering
    \resizebox{\columnwidth}{!}{
    \includegraphics[clip,trim=0 5.8in 3.8in 0.2in]{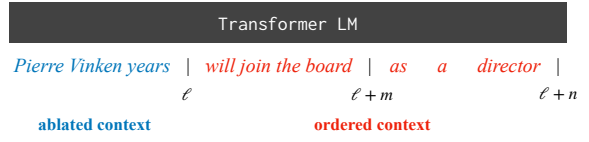}
    }
    \caption{Calculation of the ablated likelihood $\mathcal{L}(\texttt{nouns}, \ell: m\sim n)$ (\cref{eq:semi-batched}). A context ablation \texttt{nouns} (which deletes all non-noun words) is applied to the first $\ell$ tokens of the context, and likelihood is computed on the last $n-m$ (unablated) context tokens.}
    \label{fig:ablated-likelihood}
    \vspace{-1em}
\end{figure}

\textit{Batching}: 
Given a fixed (training or test) dataset of strings $\mathcal{X}$ and a \emph{maximum} context size of $m$, \cref{eq:ablated-likelihood} should be estimated empirically as $-\frac{1}{|\mathcal{X}|} \sum_x \frac{1}{|x|} \sum_{i=0}^{|x|} \log p(X_i \mid f_k(X_{i-m:i}))$. This requires re-computing model predictions once for every token in the dataset. However, the transformer models we use here support efficient \textbf{batch} inference: training data is pre-segmented into sequences of at most length $n$, and $-\frac{1}{|\mathcal{X}|n} \sum_x \sum_{i=0}^n \log p(X_i \mid f_k(X_{0:i}))$ can be computed in a single forward pass. This is considerably more efficient but means that most tokens are evaluated with a context of length $< n$. As a compromise to ensure that evaluations contain long-range context, we accumulate losses on a subset:
\begin{align}
    &\mathcal{L}(\theta, f, \ell : m \sim n) = 
     -\frac{1}{|\mathcal{X}|(n-m)} \nonumber \\[-.5em]
    &\qquad\sum_x \sum_{i=\ell+m}^{\ell+n} \log p_\theta(X_i \mid [f(X_{0:\ell}), X_{\ell:i}])
    \label{eq:semi-batched}
\end{align}
(visualized in \cref{fig:ablated-likelihood}).
This can be read as ``$\ell$ tokens of $f$-ablated context, followed by $m$ to $n$ tokens of unablated context''. We will write $\mathcal{L}(\theta, m \sim n)$ when only unablated context is used. Because of the large number of experiments in this paper, we use \cref{eq:semi-batched} for all training and evaluation.

\paragraph{Model, data and training details}

For all experiments, our LM uses the GPT-2 model architecture \cite{gpt2} in the implementation of \citet{huggingface} with default hyperparameters. All models are trained from scratch on the WikiText-103 dataset \cite{wikitext}, an English language modeling benchmark. Aside from ablations, no preprocessing is applied. A special separator token is inserted between ablated and unablated context. The training set contains 103,221,021 words, while the evaluation set contains 217,646 words.

\paragraph{A note on evaluation}

As in past work on evaluating language models \citep{brown1992estimate}, our evaluation of relative predictive information ultimately bottoms out in a conditional entropy (log-perplexity). Recent work has shown that other metrics, such as diversity of outputs, are important for evaluating the quality of LMs as models for language generation \cite{huse,gans}. Generation also depends on a number of other factors, such as choice of decoding procedure \citep{caglayan-etal-2020-curious}. Here, we focus on LMs as predictive models, measuring their ability to place an accurate distribution over future words and sentences, rather than their ability to generate useful or coherent text (see Appendix~\ref{sec:generations}). We want to emphasize that these results below apply to language models specifically, and not transformers applied to NLP tasks in general---the same analysis might give very different conclusions if applied to, e.g., question answering or summarization.

\section{Experiments}
\label{sec:experiments}

In this section, we attempt to determine what information in transformer LM contexts is usable by measuring ablated information (\cref{eq:ablated-info}). Sections \ref{sec:experiments:order} and \ref{sec:experiments:words} describe our main results, with \cref{sec:experiments:order} focused on ordering and \cref{sec:experiments:words} focused on lexical information. \cref{sec:experiments:eval} compares these results to ablations applied at evaluation time. \cref{sec:experiments:improving} explores whether contexts can be further manipulated to improve model predictions.

\subsection{Does order matter?}
\label{sec:experiments:order}

In this section we will examine the effects of different augmentations to the order within long-range context. We first train a \textbf{no information} model to minimize $\mathcal{L}(\theta, 0 \sim 512)$ and a \textbf{full information} model to minimize $\mathcal{L}(\theta, 512 \sim 1024)$. For each context ablation $f$, we train a model to minimize $\mathcal{L}(\theta, f, 512 : 0 \sim 512)$. Each ablation has access to more information than the no information model (because it conditions on extra tokens) and less information than the full information model (because an ablation has been applied to those tokens). Note that the LM operates on BPE-derived subword tokens for consistency with the way GPT-2 is typically used, but all ablations are defined at the word level, meaning, e.g., that we shuffle words rather than tokens.

\begin{figure}[t!]
    \centering
    \subfloat[\centering Mid-range condition (first 256 tokens after ablation) ]{{\includegraphics[width=0.5\textwidth]{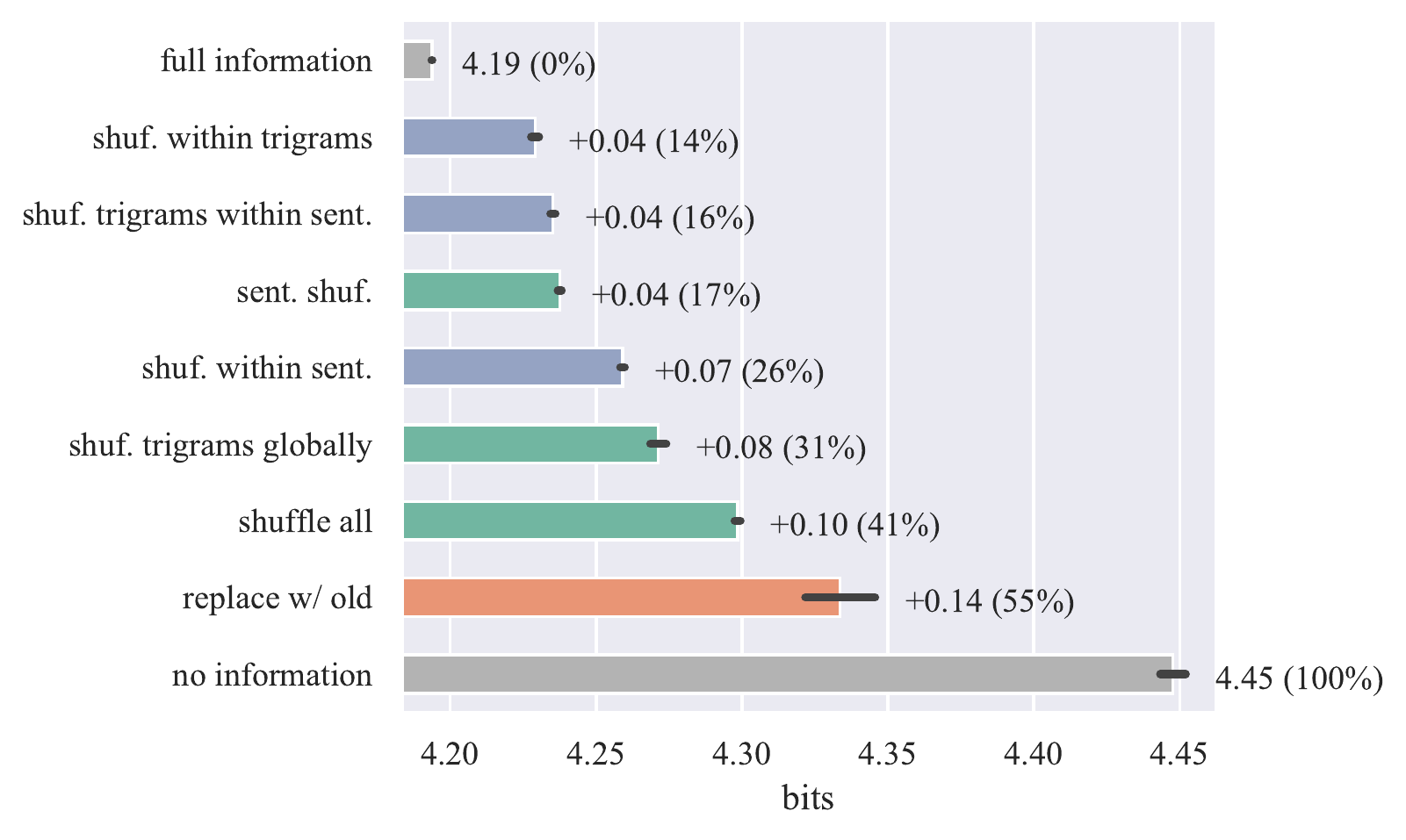} }}
    \\[-.5em]
    \hspace{-3mm}
    \subfloat[\centering Long-range condition (tokens 256-512 after ablation) ]{{\includegraphics[width=0.47\textwidth]{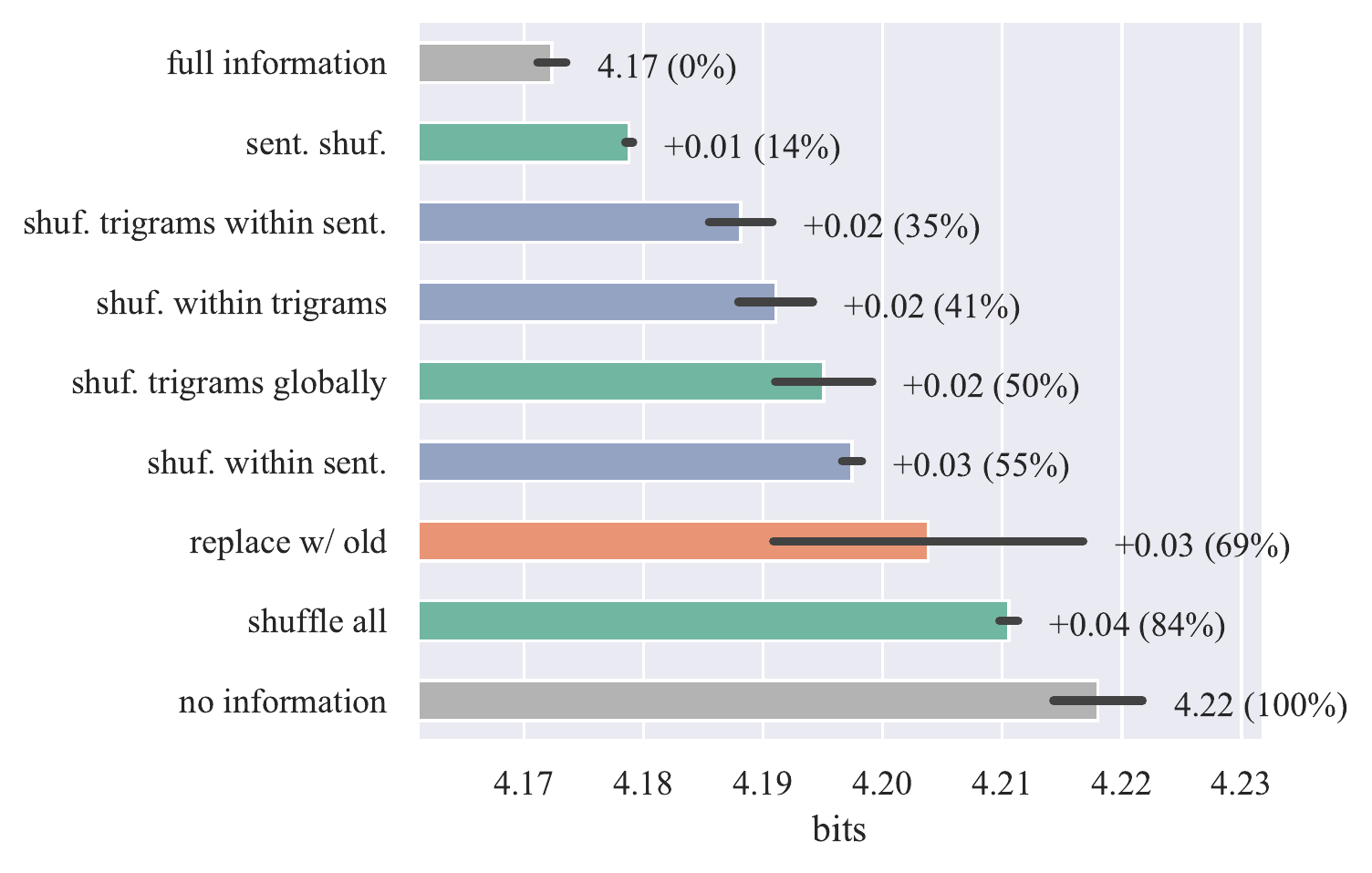} }}
    \caption{Effect of \textbf{word order} on usable information. Bar labels show ``change in ablated likelihood (ablated information)''. The $x$ axis shows ablated likelihood. Error bars represent 95\% confidence intervals. Word-order changes that preserve local ordering remove only a small amount of information, while shuffling or replacement with thematically similar text remove more.}
    \label{fig:order}
    \vspace{-1em}
\end{figure}

We use these trained models to calculate ablated information (\cref{eq:ablated-info}). To explore the effect of different context lengths, we stratify evaluation of the ablated information into two conditions: a \textbf{mid-range} condition in which likelihoods in \cref{eq:ablated-info} are of the form $\mathcal{L}(\cdot, f, 512:0 \sim 256)$, and a \textbf{long-range} condition with likelihoods $\mathcal{L}(\cdot, f, 512:256 \sim 512)$. (We call the former ``mid-range'' rather than ``short-range'' because most tokens are still predicted with significant unablated context; our experiments do not characterize sentence-internal modeling of syntactic well-formedness.) Results are shown in Figure~\ref{fig:order} and discussed below.

\paragraph{Overall word order} \strut \\[-1em]

\viztwo{shuffle all}{61 N.V., director the of Mr. Vinken Dutch group. as nonexecutive the 29. is Vinken, years Elsevier join old, publishing a Nov. will Pierre board chairman}{129}{180}{162}

\viztwo{shuf.\ trigrams globally}{publishing group. N.V., the Dutch Mr. Vinken is join the board as a nonexecutive years old, will chairman of Elsevier Pierre Vinken, 61 director Nov. 29.}{129}{180}{162}

\noindent
In the \emph{shuffle all} ablation, $f$ shuffles words uniformly at random, forcing the model to treat ablated context as a bag of words. In the \emph{shuf.\ trigrams globally} ablation, the context is divided up into non-overlapping trigrams, the order of which is then permuted uniformly at random. Shuffling all words removes 41\% of usable information in the mid-range condition and 84\% in the long-range condition: \emph{ordering information is important even very far from the target}. On the other hand, shuffling all trigrams removes 31\% of usable information in the mid-range condition and 50\% in the long-range condition: \emph{local co-occurrence statistics carry a significant amount of usable information}.

\paragraph{Word order within sentences} \strut \\[-1em]

\viztwo{shuf.\ within sent.}{61 director as the old, join will a Nov. board nonexecutive years Vinken, 29. Pierre is publishing the Vinken N.V., Mr. group. chairman Elsevier of Dutch}{151}{163}{192}

\viztwo{shuf.\ within trigrams}{Vinken, Pierre 61 will old, years the board join a nonexecutive as Nov. director 29. Mr. Vinken is of Elsevier chairman the Dutch N.V., group. publishing}{151}{163}{192}

\viztwo{shuf.\ trigrams within sent.}{years old, will as a nonexecutive join the board Pierre Vinken, 61 director Nov. 29. N.V., the Dutch chairman of Elsevier Mr. Vinken is publishing group.}{151}{163}{192}

\noindent
Words are shuffled only within sentences according to one of three procedures: (1) a uniform random permutation of all the words in the sentence (\emph{shuf.\ within sent.}), (2) a uniform random permutation of the words within each non-overlapping trigram in the sentence (\emph{shuf.\ within trigrams}), and (3) a uniform random permutation of the order of the trigrams within the sentence (\emph{shuf.\ trigrams within sent.}). (1) and (2) were also recently explored by \citet{pham2020out} in models for entailment, and more complex shuffling procedures have been explored in neuroscience contexts \citep{mollica}. Here, (2) and (3) are chosen because they preserve local co-occurrence statistics ((3) more than (2)), while (2) also preserves the general linear information flow of the sentence.

Notably, the shuf.\ within trigrams (14\% and 41\%) and the shuf.\ trigrams within sent. (16\% and 35\%) ablations both remove relatively little usable information in both the mid- and long-range conditions. \emph{Usable information is decreased only slightly by ablations that preserve local co-occurrence statistics and/or linear information flow}. (This includes transformations like \emph{man bites dog} $\to$ \emph{dog bites man} with significant effects on semantics!) In the long-range condition, uniform shuffling within sentences produces a larger effect, removing 55\% of usable information.

\paragraph{Sentence order} \strut \\[-1em]

\viztwo{shuf.\ sent.}{Mr. Vinken is chairman of Elsevier N.V., the Dutch publishing group. Pierre Vinken, 61 years old, will join the board as a nonexecutive director Nov. 29.}{129}{180}{162}

\noindent
Next, \emph{sentences} are shuffled within the context while their internal word order is unchanged. In the mid-range condition, this produces results comparable to the trigram shuffling experiments above (removing 17\% of usable information); in the long-range condition, it has an even smaller effect (14\%). Together with the previous experiment these results suggest that \emph{prediction accuracy depends on information about local word co-occurrence, but not fine-grained word order or global position}.

\paragraph{Order of entire sections} \strut \\[-1em]

\viztwo{replace w/ old}{
Rudolph Agnew, 55 years old and former chairman of Consolidated Gold Fields PLC, was named a nonexecutive director of this British industrial conglomerate.}{221}{153}{123}

\noindent
A possible hypothesis about LM behavior is that the main function of long-range context is to provide more information about the general \emph{topic} of the document, including clues about vocabulary and style. To test this, the ablation replaces its entire input with the 512 tokens that immediately precede it in the source document (which in general will be topically similar). This transformation removes significant information in both mid- and long-range conditions (55\% and 69\%). \emph{Long-range context is not simply a source of topic information: earlier text on the same theme is in some cases nearly as uninformative as no text at all.}

\begin{figure}[t!]
    \centering
    \subfloat[\centering Mid-range condition (first 256 tokens after context) ]{{\includegraphics[width=0.5\textwidth]{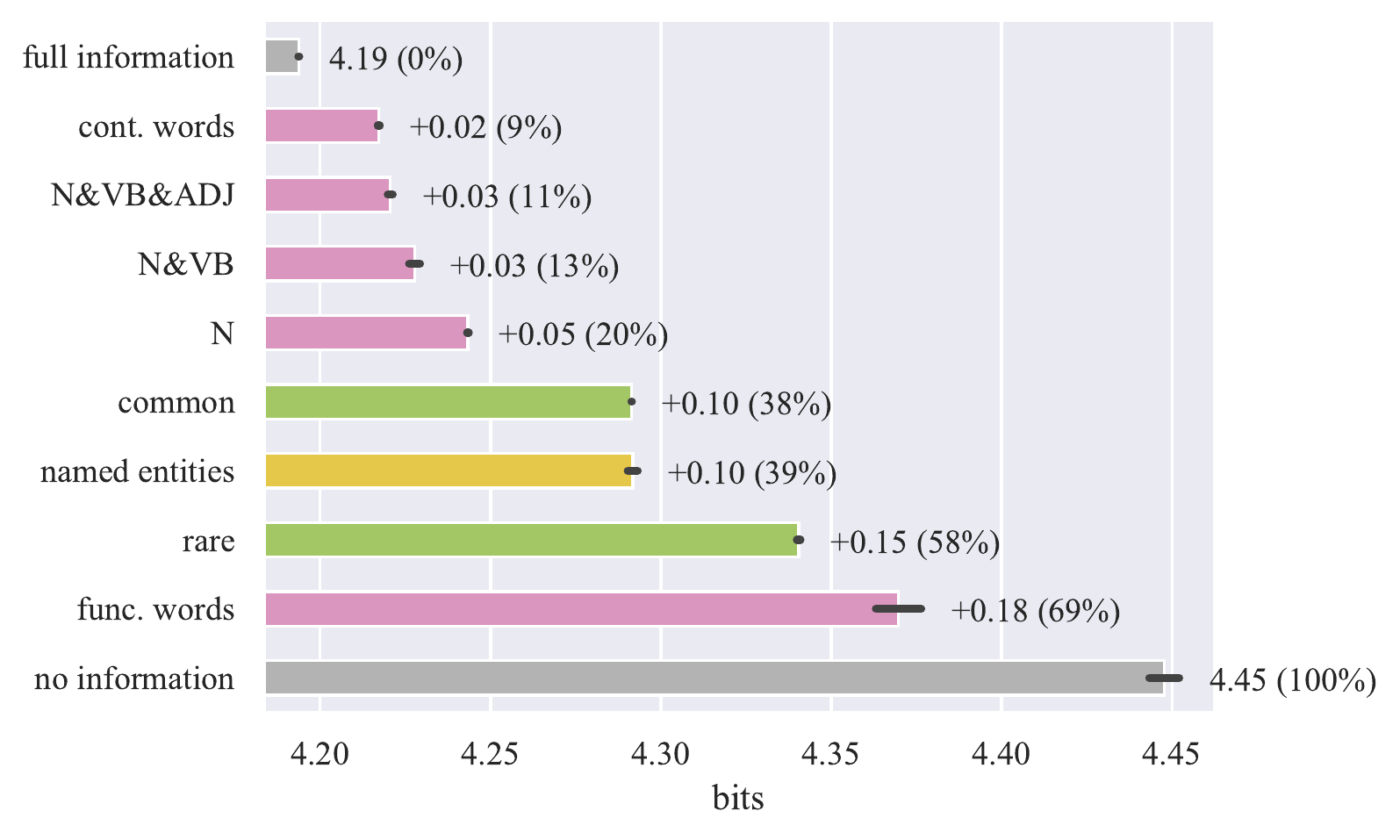} }}
    \\
    \hspace{-2mm}
    \subfloat[\centering Long-range condition (tokens 256-512 after context) ]{{\includegraphics[width=0.47\textwidth]{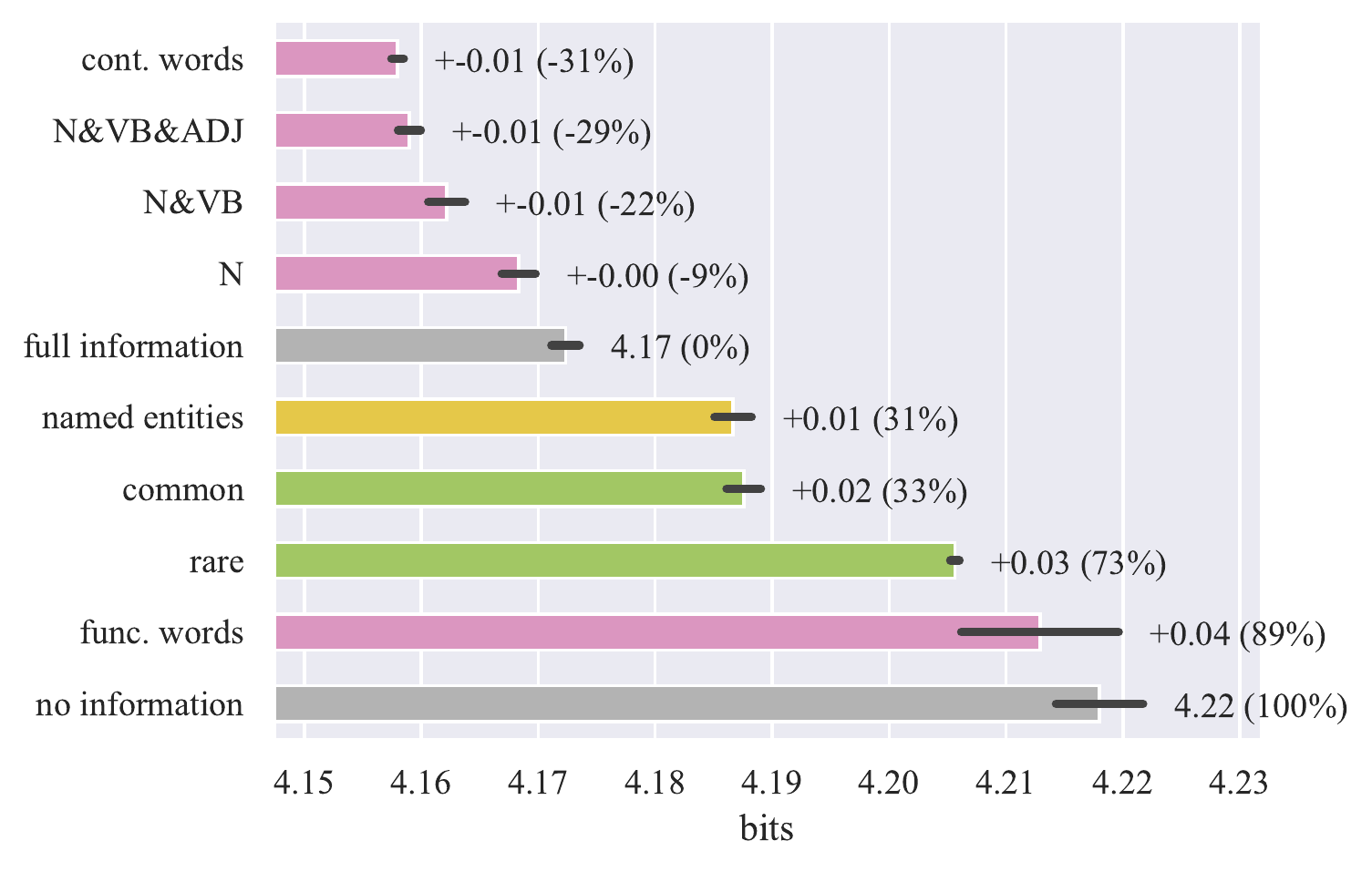} }}
    \caption{Effect of \textbf{word identity} on usable information. Labels are as in \cref{fig:order}. Several ablations, including deletion of all words except nouns, preserve most usable information in the mid-range condition, and \emph{improve} model accuracy in the in the long range.}
    \label{fig:word-class}
\end{figure}

\subsection{Do all words matter?}
\label{sec:experiments:words}

Our next experiments focus on lexical rather than structural information, using ablations that delete selected words from the context. Training and evaluation setups are exactly as in \cref{sec:experiments:order}. Here, unlike the previous section, ablations will generally cause the number of tokens in a given context to decrease; in this case ablations also insert padding tokens to the \emph{beginning} of the context window to preserve the original number of tokens. Results are shown in \cref{fig:word-class}.

\paragraph{Parts of speech} \strut \\[-1em]

\viztwo{N}{Pierre Vinken years board director Nov. Mr. Vinken chairman Elsevier N.V. publishing group}{209}{153}{190}

\viztwo{N \& VB}{Pierre Vinken years will join board director Nov. Mr. Vinken chairman Elsevier N.V. publishing group}{209}{153}{190}

\viztwo{N \& VB \& ADJ}{Pierre Vinken years old will join board nonexecutive director Nov. Mr. Vinken chairman Elsevier N.V. Dutch publishing group}{209}{153}{190}

\viztwo{cont.\ words (N \& VB \& ADJ \& ADV)}{Pierre Vinken years old will join board nonexecutive director Nov. Mr. Vinken chairman Elsevier N.V. Dutch publishing group}{209}{153}{190}

\viztwo{func.\ words}{, 61 , the as a 29 . is of , the .}{209}{153}{190}

As in the initial example from \cref{sec:approach}, we retain only words whose part of speech tag is in a given set. We use the spaCy model \cite{spacy} for part-of-speech tagging, and examine five sets: (1) \emph{nouns only}, (2) \emph{nouns and verbs}, (3) \emph{nouns, verbs, and adjectives}, (4) \emph{content words} (nouns, verbs, adjectives, and adverbs), and (5) \emph{function words} (all words \emph{except} nouns, verbs, adjectives, and adverbs).

In the mid-range condition, deleting all words but nouns removes only 20\% of usable information; deleting all but nouns and verbs removes only 13\%. \emph{Most usable information, even in mid-range context, appears to be captured by nouns and verbs.} Retaining only function words causes a considerably greater loss of information.

In the long-range condition, results are even more striking: \emph{retaining only content words \emph{improves} predictions over the ``full information'' experiment}. Like Shannon information, $\V$-information is defined to be non-negative \cite{usable-information}, and the result in \cref{fig:word-class} is a consequence of our finite-sample approximation based on held-out likelihood. The effect is robust across multiple training runs from random initializations. As there is a significant gap between the training and validation perplexity of our model (roughly 11\%), we hypothesize that this change occurs because the ablation preserves semantic content while reducing the original model's ability to overfit. We believe this is an important subject for future investigation.

\paragraph{Named entities} \strut \\[-1em]

\viztwo{named entities}{Pierre Vinken 61 years old Nov. 29 Vinken Elsevier N.V. Dutch}{225}{200}{96}

\noindent
As an alternative to the \emph{topic hypothesis} evaluated under ``Order of entire sections'' above, we might hypothesize that long-range contexts are useful because they provide a reservoir of named entities likely to be referred to again. Here, the ablation retains only spans tagged as named entities or quantities by spaCy. While significantly worse than the noun ablation discussed above, retaining only entities results removes only about a third of usable information in both conditions (39\% and 31\%).

\paragraph{Word frequency} \strut \\[-1em]

\viztwo{common}{Pierre years old join board director . Mr. chairman Dutch publishing group .}{170}{197}{113}

\viztwo{rare}{Vinken nonexecutive Nov. Vinken Elsevier N.V.}{170}{197}{113}

\noindent
Another natural question is whether rare words or frequent words are more important: information about frequent context words might help models estimate fine-grained document-level frequencies of those words account for most of the terms in \cref{eq:ablated-likelihood}; rare words are likely to be more informative about the content of the document itself.

We partition the vocabulary into a set of \emph{rare words}, corresponding to the least frequent $\sim 98\%$ of word types and 20\% of word tokens, and \emph{frequent words}, the most frequent $\sim 2\%$ of types and 80\% of tokens. Both ablations remove a significant amount of information relative to the POS-based ablations above, but retaining only frequent words improves perplexity relative to rare words in both the mid- and long-range conditions.

Appendix~\ref{sec:longer-context} presents versions of these experiments trained and evaluated on even longer contexts. Conclusions are largely the same as above.

\begin{figure}[t!]
    \centering
    \subfloat[\centering Mid-range condition (first 256 tokens after ablation) ]{{\includegraphics[width=0.5\textwidth]{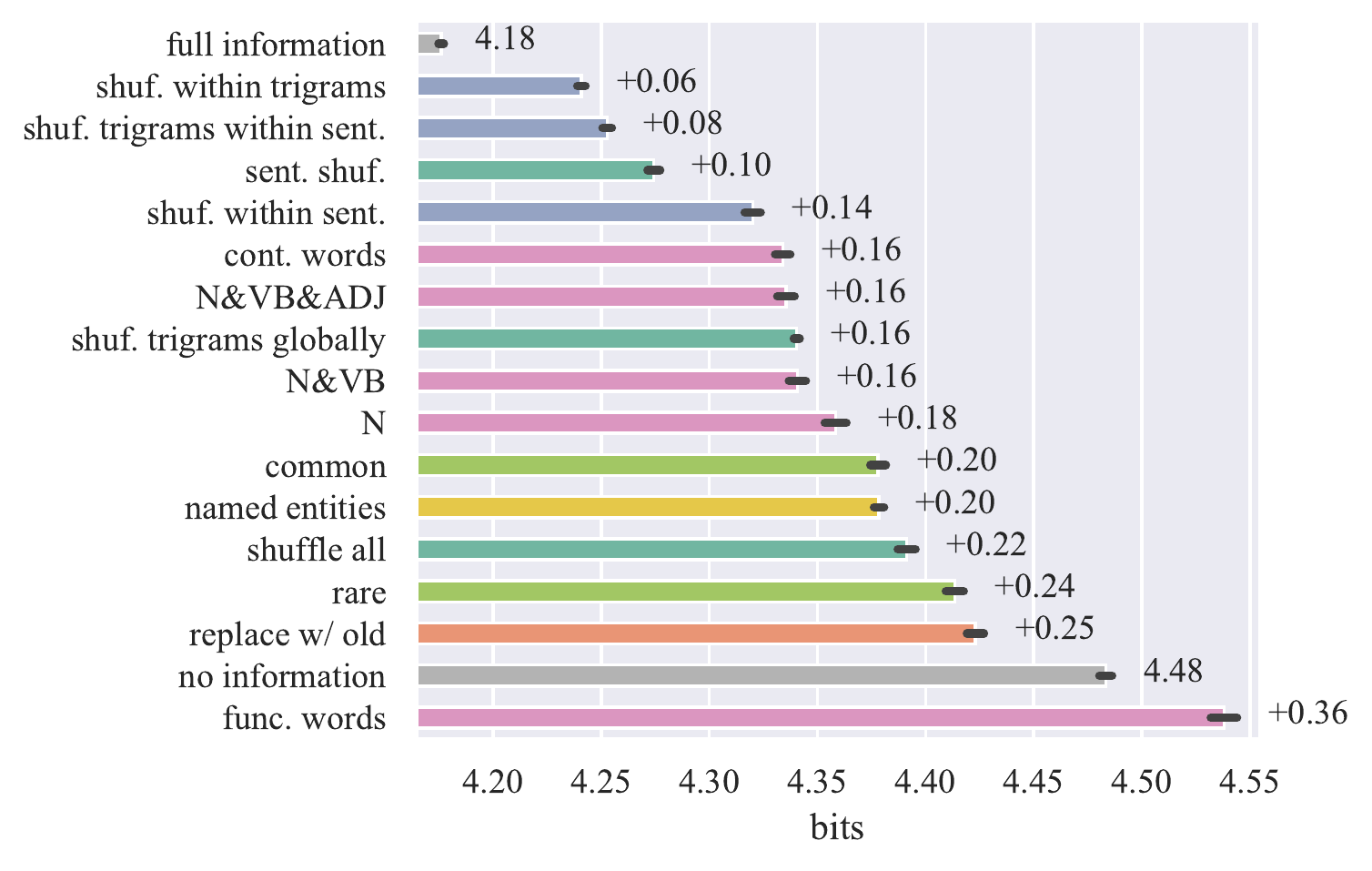} }}
    \\
    \subfloat[\centering Long-range condition (tokens 256-512 after ablation) ]{{\includegraphics[width=0.47\textwidth]{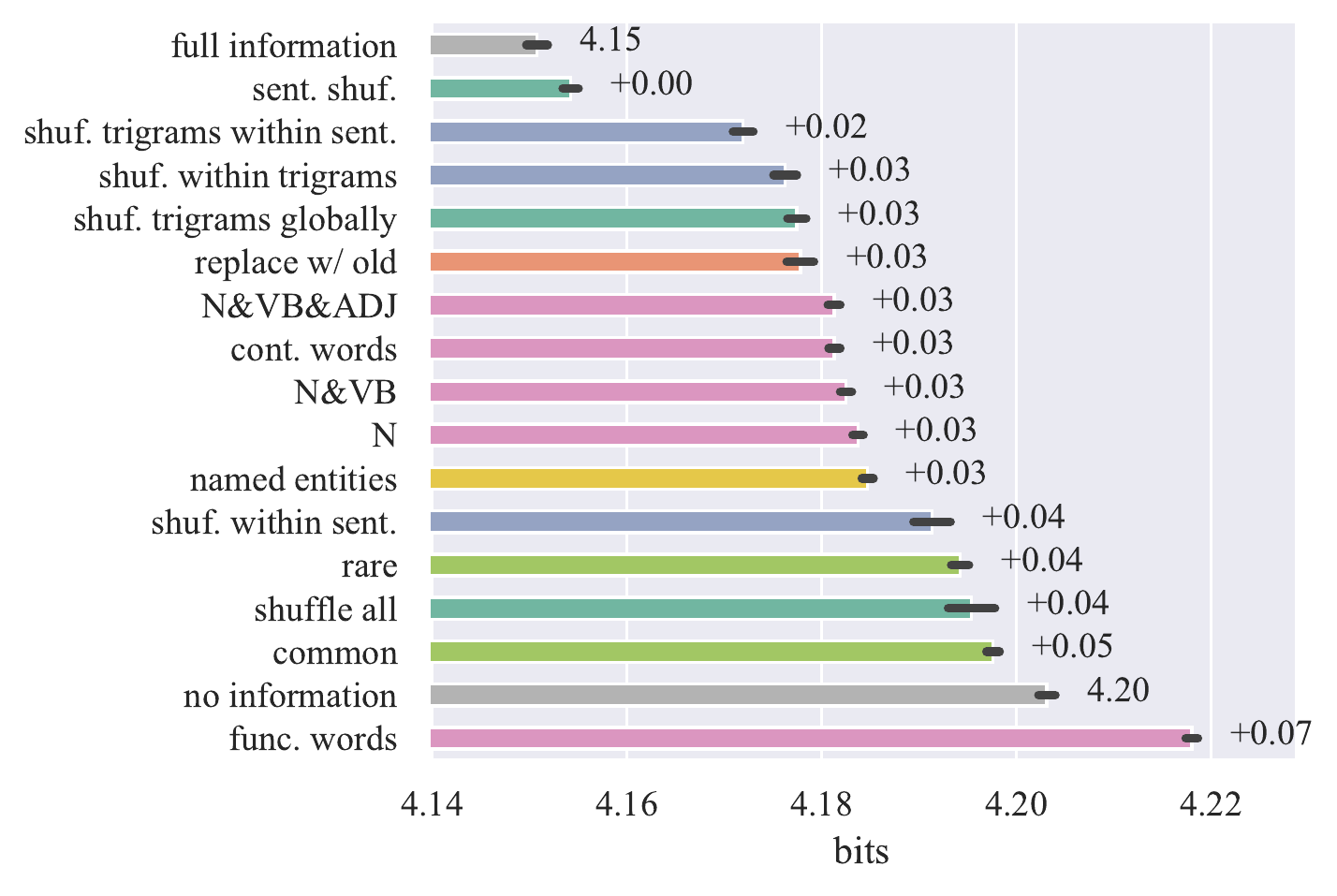} }}
    \caption{Loss of information resulting from ablations at \emph{evaluation time only}. $x$-axis and labels show ablated negative log-likelihoods. Some locality-preserving ablations (high PMI, shuf.\ sent.) have a small effect, but most affect likelihood significantly (including lexical ablations that do not remove usable information).}
    \label{fig:eval}
    \vspace{-1em}
\end{figure}

\subsection{Evaluating on augmented data}
\label{sec:experiments:eval}

We motivated the use of $\V$-information in \cref{sec:approach} by arguing that it more clearly distinguished between prediction errors attributable to \emph{loss of information} and prediction errors attributable to \emph{malformed and out-of-distribution} model inputs. To put our results in context, we repeat several of the previous experiments in the evaluation paradigm of \citet{sharp-fuzzy}, which is designed to measure test-time sensitivity rather than usable information.

We train a new model to minimize $\mathcal{L}(\theta, 512 \sim 1024)$ while randomly truncating the first 512 context tokens and replacing them with padding tokens (to ensure that the model has seen padding tokens at training time). We then evaluate this model on the set of ablations shown in \cref{sec:experiments:order} and \cref{sec:experiments:words}. For the full information model in \cref{fig:eval}, we evaluate on ordered context windows with no padding tokens; for the no information model, we evaluate on context windows in which the first 512 tokens are all padding tokens.

In the mid-range condition, the least destructive ablations are shuffling within trigrams and shuffling the order of trigrams within sentences: models appear to be reasonably robust to this kind of data transformation without specific training on it. Importantly, lexical ablation experiments have a large impact in this evaluation, underlining the extent to which the two experimental paradigms characterize different aspects of model behavior. Figure~\ref{fig:compare} in Appendix~\ref{sec:comparison} shows a side-by-side comparison of these experiments and the ones in Sections~\ref{sec:experiments:order}--\ref{sec:experiments:words}.

\subsection{Making better language models?}
\label{sec:experiments:improving}
The lexical ablation experiments in \cref{sec:experiments:words} indicated that model accuracy could be \emph{improved} by selective deletion of context words. Can this effect be exploited to further improve models? As a simple experiment, we attempted to \emph{replace} all padding tokens in the \emph{nouns+verbs} ablation of \cref{sec:experiments:words} with nouns and verbs from further back in the context---effectively providing the model with an even longer-range view of an informative context representation.

This experiment slightly increased usable information in the mid-range condition (0.2\%), but \emph{decreased} it in the long range-range condition (0.6\%). \emph{Longer contexts, even of a kind previously found to be informative, did not provide additional usable information.} These results are consistent with our earlier hypothesis that the previously observed effect resulted from a reduction in overfitting---if removing information increased performance by reducing overfitting, then it is reasonable that adding information back results in more overfitting.

\section{Related Work}

\paragraph{Context in count-based and discriminative LMs}

The earliest learned LMs were \emph{count-based} \citep[e.g.,][]{kneser1995improved}: they estimated $p(x_n \mid x_{0:n})$ based on a (smoothed) empirical $n$-gram frequency $\#(x_{0:n}) / \#(x_{0:n-1})$ (where $\#(x)$ is the number of times the sequence $x$ appears in training data). As the number of distinct $n$-gram counts grows exponentially in $n$, it was typically set to a small value. Count-based models have a clear dependence on context: any token within the last $n$ words that also appears in a training n-gram is relevant, anything further back is not.

Subsequent models improved on these by allowing the use of skip-grams, caches, and feature-based models \citep{goodman2001bit, bengio2003neural}. Some of these in principle allowed the use of unlimited-length contexts, but only by imposing strong restrictions on the ways in which context features could interact.

\paragraph{Context in RNN LMs}

Recurrent neural network language models \citep{mikolov2010recurrent, elman1990finding} provide a more expressive mechanism for the use of long-range context: models write to a recurrent ``state vector'' which can be carried arbitrarily far into the future. Computational issues limit the effective context size such models can be practically trained on, but this size is still significantly greater the models mentioned above: as previously noted, \citet{sharp-fuzzy} revealed influence from up to 200 tokens of context. Similar effects are reported by \citet{sankar2019neural} for neural dialogue models, and \citet{li2016understanding} describe an alternative procedure for ablating contexts.

\paragraph{Context in Transformer LMs}

Transformers introduce yet another mechanism for extracting information from long-range context: attention. Attention is also used with RNNs, but typically with just a single head---the hidden state still carries most of the information. In transformers, context enters into predictions primarily via unbounded random access. These models appear to benefit from significantly longer contexts than previous models.

Some recent work that investigates the behavior of individual transformer attention heads \cite{bert-analysis,specialized-heads}. This work finds that certain attention heads are sensitive to things like word frequency, positional information, and certain syntactic phenomena. While extremely informative about the computational structures implemented by fixed models, these approaches do not necessarily reveal anything about usable information: indeed, patterns of attention do not necessarily correlate with model predictions \cite{attention-neq-explanation}.

\paragraph{Other related work}

Our finding that fine-grained ordering information contributes little usable information is consistent with \citet{compressive}'s finding that long-range contexts could be informatively summarized in fixed-sized vectors; our finding that most usable information is carried by nouns is consistent with earlier findings about both specialized neural architectures \citep{henaff2016tracking} and discourse representations in feature-based models \citep{barzilay2008modeling}. Our approach also shares similar motivations to information-theoretic work on \emph{probing} \citep{voita-titov-2020-information, pimentel-etal-2020-information}, which uses related tools to interpret linguistic structure in LM representations rather than characterizing their effect on LM predictions. Several recent papers have explored the effect of training-time and test-time ablations in models for other data analysis tasks: \citet{pham2020out} find that shuffling experiments have a limited effect on the accuracy of models for natural language inference, while \citet{perez2021rissanen} describe several experiments aimed at \emph{introducing} usable information for several question answering and sentence understanding tasks.

\section{Discussion}
\label{sec:discussion}

We have investigated the extent to which transformer models can use structural and lexical information in long-range contexts for English language modeling. Experiments demonstrated that this information is primarily contained in content words and local ordering statistics: ablations that remove other kinds of information from context have little effect on models' predictive accuracies. In contrast, retaining only information about document identity or named entities causes significant drops in predictive accuracy: the effectiveness of long contexts is not explained by the presence of topic or named entity information alone. 

Crucial to obtaining these results was a measure of \emph{ablated usable information} grounded in the accuracy of models trained and tested on ablated contexts. Past work on context in LMs has primarily measured the influence of evaluation-time ablations. Sometimes these two notions of context-sensitivity coincide (e.g., trigram shuffling) and sometimes they do not (e.g., removal of lexical information). Our results also offer a jumping-off point for future modeling work. They motivate more \emph{efficient}, compressed context representations that better preserve the information that is usable by current models. They motivate more \emph{accurate} models by developing new context representations that make currently unusable information more prominent.

Several questions remain unanswered by our experiments. Do ablations affect the quality of text generated by models? (In particular, does the usable information added by long contexts improve predictability of syntax, semantics, or simply document-level word frequency statistics?) More fundamentally, do observations about usable information reflect limitations of transformers or fundamental, (Shannon-)information-theoretic properties of English? Our results suggest that at least some of these effects are model-specific: deleting function words cannot add information, but improves held-out model accuracy. A complete answer to this question will require more detailed exploration, including a better understanding of human predictions in comparable settings.

\section*{Acknowledgments}

Thanks to Carina Kauf and Greta Tuckute, Evelina Fedorenko and Roger Levy for valuable discussions. We acknowledge the MIT SuperCloud and Lincoln Laboratory Supercomputing Center for providing HPC resources that contributed to the results reported within this paper.

\section*{Impact Statement}

Across initial exploration, evaluation conditions and training runs, experiments in this paper required roughly 100 training runs on the WikiText-103 dataset. As discussed in \cref{sec:approach}, model size and batched evaluation were both used to minimize the energy demands of these experiments; experiments themselves were performed at the Massachusetts Green HPC center, a carbon-neutral supercomputing facility. Ultimately, results in \cref{sec:experiments} provide guidance toward the design of models that use context more efficiently and motivate the large-scale empirical study conducted here.

\bibliographystyle{acl_natbib}
\bibliography{anthology,acl2021}

\begin{thebibliography}{39}
\expandafter\ifx\csname natexlab\endcsname\relax\def\natexlab#1{#1}\fi

\bibitem[{Barzilay and Lapata(2008)}]{barzilay2008modeling}
Regina Barzilay and Mirella Lapata. 2008.
\newblock Modeling local coherence: An entity-based approach.
\newblock \emph{Computational Linguistics}, 34(1):1--34.

\bibitem[{Beltagy et~al.(2020)Beltagy, Peters, and Cohan}]{longformer}
Iz~Beltagy, Matthew~E. Peters, and Arman Cohan. 2020.
\newblock Longformer: The long-document transformer.
\newblock \emph{arXiv:2004.05150}.

\bibitem[{Bengio et~al.(2003)Bengio, Ducharme, Vincent, and
  Janvin}]{bengio2003neural}
Yoshua Bengio, R{\'e}jean Ducharme, Pascal Vincent, and Christian Janvin. 2003.
\newblock A neural probabilistic language model.
\newblock \emph{The Journal of Machine Learning Research}, 3:1137--1155.

\bibitem[{Brown et~al.(1992)Brown, Della~Pietra, Della~Pietra, Lai, and
  Mercer}]{brown1992estimate}
Peter~F Brown, Stephen~A Della~Pietra, Vincent~J Della~Pietra, Jennifer~C Lai,
  and Robert~L Mercer. 1992.
\newblock An estimate of an upper bound for the entropy of english.
\newblock \emph{Computational Linguistics}, 18(1):31--40.

\bibitem[{Brown(2011)}]{brown2011cmu}
Ralf~D Brown. 2011.
\newblock The {CMU-EBMT} machine translation system.
\newblock \emph{Machine translation}, 25(2):179.

\bibitem[{Caccia et~al.(2020)Caccia, Caccia, Fedus, Larochelle, Pineau, and
  Charlin}]{gans}
Massimo Caccia, Lucas Caccia, William Fedus, Hugo Larochelle, Joelle Pineau,
  and Laurent Charlin. 2020.
\newblock \href {https://openreview.net/forum?id=BJgza6VtPB} {Language {GANs}
  falling short}.
\newblock In \emph{International Conference on Learning Representations}.

\bibitem[{Caglayan et~al.(2020)Caglayan, Madhyastha, and
  Specia}]{caglayan-etal-2020-curious}
Ozan Caglayan, Pranava Madhyastha, and Lucia Specia. 2020.
\newblock \href {https://www.aclweb.org/anthology/2020.coling-main.210}
  {Curious case of language generation evaluation metrics: A cautionary tale}.
\newblock In \emph{Proceedings of the 28th International Conference on
  Computational Linguistics}, pages 2322--2328, Barcelona, Spain (Online).
  International Committee on Computational Linguistics.

\bibitem[{Child et~al.(2019)Child, Gray, Radford, and Sutskever}]{sparse}
Rewon Child, Scott Gray, Alec Radford, and Ilya Sutskever. 2019.
\newblock Generating long sequences with sparse transformers.
\newblock \emph{URL https://openai.com/blog/sparse-transformers}.

\bibitem[{Clark et~al.(2019)Clark, Khandelwal, Levy, and
  Manning}]{bert-analysis}
Kevin Clark, Urvashi Khandelwal, Omer Levy, and Christopher~D. Manning. 2019.
\newblock \href {https://doi.org/10.18653/v1/W19-4828} {What does {BERT} look
  at? an analysis of {BERT}{'}s attention}.
\newblock In \emph{Proceedings of the 2019 ACL Workshop BlackboxNLP: Analyzing
  and Interpreting Neural Networks for NLP}, pages 276--286, Florence, Italy.
  Association for Computational Linguistics.

\bibitem[{Dai et~al.(2019)Dai, Yang, Yang, Carbonell, Le, and
  Salakhutdinov}]{transformer-xl}
Zihang Dai, Zhilin Yang, Yiming Yang, Jaime Carbonell, Quoc Le, and Ruslan
  Salakhutdinov. 2019.
\newblock \href {https://doi.org/10.18653/v1/P19-1285} {Transformer-{XL}:
  Attentive language models beyond a fixed-length context}.
\newblock In \emph{Proceedings of the 57th Annual Meeting of the Association
  for Computational Linguistics}, pages 2978--2988, Florence, Italy.
  Association for Computational Linguistics.

\bibitem[{Elman(1990)}]{elman1990finding}
Jeffrey~L Elman. 1990.
\newblock Finding structure in time.
\newblock \emph{Cognitive science}, 14(2):179--211.

\bibitem[{Goodman(2001)}]{goodman2001bit}
Joshua~T Goodman. 2001.
\newblock A bit of progress in language modeling.
\newblock \emph{Computer Speech \& Language}, 15(4):403--434.

\bibitem[{Hahn(2020)}]{hahn-2020-theoretical}
Michael Hahn. 2020.
\newblock \href {https://doi.org/10.1162/tacl_a_00306} {Theoretical limitations
  of self-attention in neural sequence models}.
\newblock \emph{Transactions of the Association for Computational Linguistics},
  8:156--171.

\bibitem[{Hashimoto et~al.(2019)Hashimoto, Zhang, and Liang}]{huse}
Tatsunori Hashimoto, Hugh Zhang, and Percy Liang. 2019.
\newblock \href {https://doi.org/10.18653/v1/N19-1169} {Unifying human and
  statistical evaluation for natural language generation}.
\newblock In \emph{Proceedings of the 2019 Conference of the North {A}merican
  Chapter of the Association for Computational Linguistics: Human Language
  Technologies, Volume 1 (Long and Short Papers)}, pages 1689--1701,
  Minneapolis, Minnesota. Association for Computational Linguistics.

\bibitem[{Henaff et~al.(2016)Henaff, Weston, Szlam, Bordes, and
  LeCun}]{henaff2016tracking}
Mikael Henaff, Jason Weston, Arthur Szlam, Antoine Bordes, and Yann LeCun.
  2016.
\newblock Tracking the world state with recurrent entity networks.
\newblock In \emph{ICLR}.

\bibitem[{Honnibal et~al.(2020)Honnibal, Montani, Van~Landeghem, and
  Boyd}]{spacy}
Matthew Honnibal, Ines Montani, Sofie Van~Landeghem, and Adriane Boyd. 2020.
\newblock \href {https://doi.org/10.5281/zenodo.1212303} {{spaCy:
  Industrial-strength Natural Language Processing in Python}}.

\bibitem[{Jain and Wallace(2019)}]{attention-neq-explanation}
Sarthak Jain and Byron~C. Wallace. 2019.
\newblock Attention is not explanation.
\newblock In \emph{NAACL-HLT}.

\bibitem[{Khandelwal et~al.(2018)Khandelwal, He, Qi, and
  Jurafsky}]{sharp-fuzzy}
Urvashi Khandelwal, He~He, Peng Qi, and Dan Jurafsky. 2018.
\newblock \href {https://doi.org/10.18653/v1/P18-1027} {Sharp nearby, fuzzy far
  away: How neural language models use context}.
\newblock In \emph{Proceedings of the 56th Annual Meeting of the Association
  for Computational Linguistics (Volume 1: Long Papers)}, pages 284--294,
  Melbourne, Australia. Association for Computational Linguistics.

\bibitem[{Kitaev et~al.(2020)Kitaev, Kaiser, and Levskaya}]{reformer}
Nikita Kitaev, Lukasz Kaiser, and Anselm Levskaya. 2020.
\newblock \href {https://openreview.net/forum?id=rkgNKkHtvB} {Reformer: The
  efficient transformer}.
\newblock In \emph{International Conference on Learning Representations}.

\bibitem[{Kneser and Ney(1995)}]{kneser1995improved}
Reinhard Kneser and Hermann Ney. 1995.
\newblock Improved backing-off for m-gram language modeling.
\newblock In \emph{1995 International Conference on Acoustics, Speech, and
  Signal Processing}, volume~1, pages 181--184. IEEE.

\bibitem[{Li et~al.(2016)Li, Monroe, and Jurafsky}]{li2016understanding}
Jiwei Li, Will Monroe, and Dan Jurafsky. 2016.
\newblock Understanding neural networks through representation erasure.
\newblock \emph{arXiv preprint arXiv:1612.08220}.

\bibitem[{Merity et~al.(2016)Merity, Xiong, Bradbury, and Socher}]{wikitext}
Stephen Merity, Caiming Xiong, James Bradbury, and Richard Socher. 2016.
\newblock \href {http://arxiv.org/abs/1609.07843} {Pointer sentinel mixture
  models}.
\newblock \emph{CoRR}, abs/1609.07843.

\bibitem[{Mikolov et~al.(2010)Mikolov, Karafi{\'a}t, Burget,
  {\v{C}}ernock{\`y}, and Khudanpur}]{mikolov2010recurrent}
Tom{\'a}{\v{s}} Mikolov, Martin Karafi{\'a}t, Luk{\'a}{\v{s}} Burget, Jan
  {\v{C}}ernock{\`y}, and Sanjeev Khudanpur. 2010.
\newblock Recurrent neural network based language model.
\newblock In \emph{Eleventh annual conference of the International Speech
  Communication Association}.

\bibitem[{Mollica et~al.(2020)Mollica, Siegelman, Diachek, Piantadosi,
  Mineroff, Futrell, Kean, Qian, and Fedorenko}]{mollica}
F.~Mollica, Matthew Siegelman, Evgeniia Diachek, S.~Piantadosi, Zachary
  Mineroff, Richard Futrell, Hope~H. Kean, Peng Qian, and E.~Fedorenko. 2020.
\newblock Composition is the core driver of the language-selective network.
\newblock \emph{Neurobiology of Language}, 1:104--134.

\bibitem[{Perez et~al.(2021)Perez, Kiela, and Cho}]{perez2021rissanen}
Ethan Perez, Douwe Kiela, and Kyunghyun Cho. 2021.
\newblock Rissanen data analysis: Examining dataset characteristics via
  description length.
\newblock \emph{arXiv preprint arXiv:2103.03872}.

\bibitem[{Pham et~al.(2020)Pham, Bui, Mai, and Nguyen}]{pham2020out}
Thang~M Pham, Trung Bui, Long Mai, and Anh Nguyen. 2020.
\newblock Out of order: How important is the sequential order of words in a
  sentence in natural language understanding tasks?
\newblock \emph{arXiv preprint arXiv:2012.15180}.

\bibitem[{Pimentel et~al.(2020)Pimentel, Valvoda, Hall~Maudslay, Zmigrod,
  Williams, and Cotterell}]{pimentel-etal-2020-information}
Tiago Pimentel, Josef Valvoda, Rowan Hall~Maudslay, Ran Zmigrod, Adina
  Williams, and Ryan Cotterell. 2020.
\newblock \href {https://doi.org/10.18653/v1/2020.acl-main.420}
  {Information-theoretic probing for linguistic structure}.
\newblock In \emph{Proceedings of the 58th Annual Meeting of the Association
  for Computational Linguistics}, pages 4609--4622, Online. Association for
  Computational Linguistics.

\bibitem[{Radford et~al.(2019)Radford, Wu, Child, Luan, Amodei, and
  Sutskever}]{gpt2}
Alec Radford, Jeff Wu, Rewon Child, David Luan, Dario Amodei, and Ilya
  Sutskever. 2019.
\newblock Language models are unsupervised multitask learners.

\bibitem[{Rae et~al.(2019)Rae, Potapenko, Jayakumar, and
  Lillicrap}]{compressive}
Jack~W. Rae, Anna Potapenko, Siddhant~M. Jayakumar, and Timothy~P. Lillicrap.
  2019.
\newblock \href {http://arxiv.org/abs/1911.05507} {Compressive transformers for
  long-range sequence modelling}.

\bibitem[{Sankar et~al.(2019)Sankar, Subramanian, Pal, Chandar, and
  Bengio}]{sankar2019neural}
Chinnadhurai Sankar, Sandeep Subramanian, Christopher Pal, Sarath Chandar, and
  Yoshua Bengio. 2019.
\newblock Do neural dialog systems use the conversation history effectively? an
  empirical study.
\newblock \emph{arXiv preprint arXiv:1906.01603}.

\bibitem[{Shannon(1948)}]{shannon1948mathematical}
Claude~E Shannon. 1948.
\newblock A mathematical theory of communication.
\newblock \emph{The Bell system technical journal}, 27(3):379--423.

\bibitem[{Vaswani et~al.(2017)Vaswani, Shazeer, Parmar, Uszkoreit, Jones,
  Gomez, Kaiser, and Polosukhin}]{transformer}
Ashish Vaswani, Noam Shazeer, Niki Parmar, Jakob Uszkoreit, Llion Jones,
  Aidan~N Gomez, \L~ukasz Kaiser, and Illia Polosukhin. 2017.
\newblock \href
  {http://papers.nips.cc/paper/7181-attention-is-all-you-need.pdf} {Attention
  is all you need}.
\newblock In I.~Guyon, U.~V. Luxburg, S.~Bengio, H.~Wallach, R.~Fergus,
  S.~Vishwanathan, and R.~Garnett, editors, \emph{Advances in Neural
  Information Processing Systems 30}, pages 5998--6008. Curran Associates, Inc.

\bibitem[{Voita et~al.(2019)Voita, Talbot, Moiseev, Sennrich, and
  Titov}]{specialized-heads}
Elena Voita, David Talbot, Fedor Moiseev, Rico Sennrich, and Ivan Titov. 2019.
\newblock \href {https://doi.org/10.18653/v1/P19-1580} {Analyzing multi-head
  self-attention: Specialized heads do the heavy lifting, the rest can be
  pruned}.
\newblock In \emph{Proceedings of the 57th Annual Meeting of the Association
  for Computational Linguistics}, pages 5797--5808, Florence, Italy.
  Association for Computational Linguistics.

\bibitem[{Voita and Titov(2020)}]{voita-titov-2020-information}
Elena Voita and Ivan Titov. 2020.
\newblock \href {https://doi.org/10.18653/v1/2020.emnlp-main.14}
  {Information-theoretic probing with minimum description length}.
\newblock In \emph{Proceedings of the 2020 Conference on Empirical Methods in
  Natural Language Processing (EMNLP)}, pages 183--196, Online. Association for
  Computational Linguistics.

\bibitem[{Wang et~al.(2019)Wang, Pruksachatkun, Nangia, Singh, Michael, Hill,
  Levy, and Bowman}]{wang2019superglue}
Alex Wang, Yada Pruksachatkun, Nikita Nangia, Amanpreet Singh, Julian Michael,
  Felix Hill, Omer Levy, and Samuel~R Bowman. 2019.
\newblock {SuperGLUE}: A stickier benchmark for general-purpose language
  understanding systems.
\newblock \emph{arXiv preprint arXiv:1905.00537}.

\bibitem[{Wang et~al.(2020)Wang, Li, Khabsa, Fang, and Ma}]{linformer}
Sinong Wang, Belinda Li, Madian Khabsa, Han Fang, and Hao Ma. 2020.
\newblock Linformer: Self-attention with linear complexity.
\newblock \emph{arXiv preprint arXiv:2006.04768}.

\bibitem[{Wolf et~al.(2020)Wolf, Debut, Sanh, Chaumond, Delangue, Moi, Cistac,
  Rault, Louf, Funtowicz, Davison, Shleifer, von Platen, Ma, Jernite, Plu, Xu,
  Scao, Gugger, Drame, Lhoest, and Rush}]{huggingface}
Thomas Wolf, Lysandre Debut, Victor Sanh, Julien Chaumond, Clement Delangue,
  Anthony Moi, Pierric Cistac, Tim Rault, Rémi Louf, Morgan Funtowicz, Joe
  Davison, Sam Shleifer, Patrick von Platen, Clara Ma, Yacine Jernite, Julien
  Plu, Canwen Xu, Teven~Le Scao, Sylvain Gugger, Mariama Drame, Quentin Lhoest,
  and Alexander~M. Rush. 2020.
\newblock \href {https://www.aclweb.org/anthology/2020.emnlp-demos.6}
  {Transformers: State-of-the-art natural language processing}.
\newblock In \emph{Proceedings of the 2020 Conference on Empirical Methods in
  Natural Language Processing: System Demonstrations}, pages 38--45, Online.
  Association for Computational Linguistics.

\bibitem[{Xu et~al.(2020)Xu, Zhao, Song, Stewart, and
  Ermon}]{usable-information}
Yilun Xu, Shengjia Zhao, Jiaming Song, Russell Stewart, and Stefano Ermon.
  2020.
\newblock \href {https://openreview.net/forum?id=r1eBeyHFDH} {A theory of
  usable information under computational constraints}.
\newblock In \emph{International Conference on Learning Representations}.

\bibitem[{Zhang et~al.(2016)Zhang, Bengio, Hardt, Recht, and
  Vinyals}]{zhang2016understanding}
Chiyuan Zhang, Samy Bengio, Moritz Hardt, Benjamin Recht, and Oriol Vinyals.
  2016.
\newblock Understanding deep learning requires rethinking generalization.
\newblock \emph{arXiv preprint arXiv:1611.03530}.

\end{thebibliography}

\appendix

\section{Comparison of Experimental Paradigms}
\label{sec:comparison}

\begin{figure}[ht!]
    \centering
    \subfloat[\centering Mid-range condition (first 256 tokens after ablation) ]{{\includegraphics[width=0.5\textwidth]{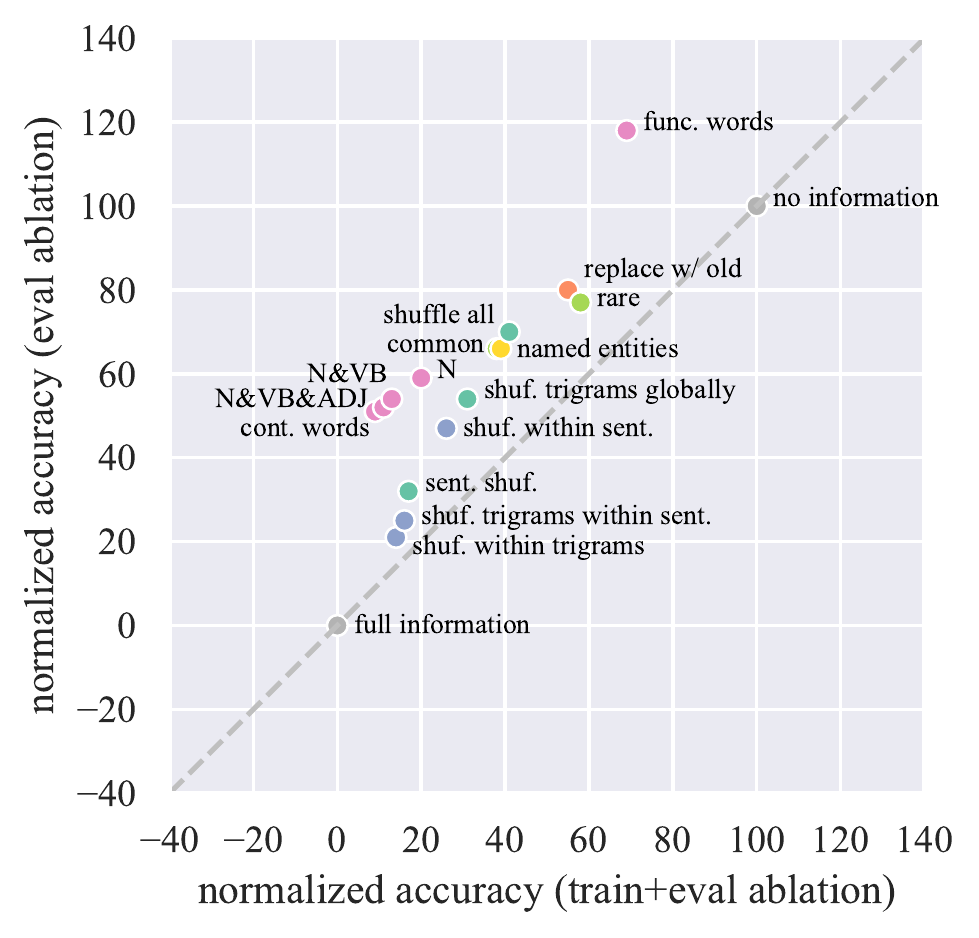} }}
    \\[-.5em]
    \hspace{-3mm}
    \subfloat[\centering Long-range condition (tokens 256-512 after ablation) ]{{\includegraphics[width=0.5\textwidth]{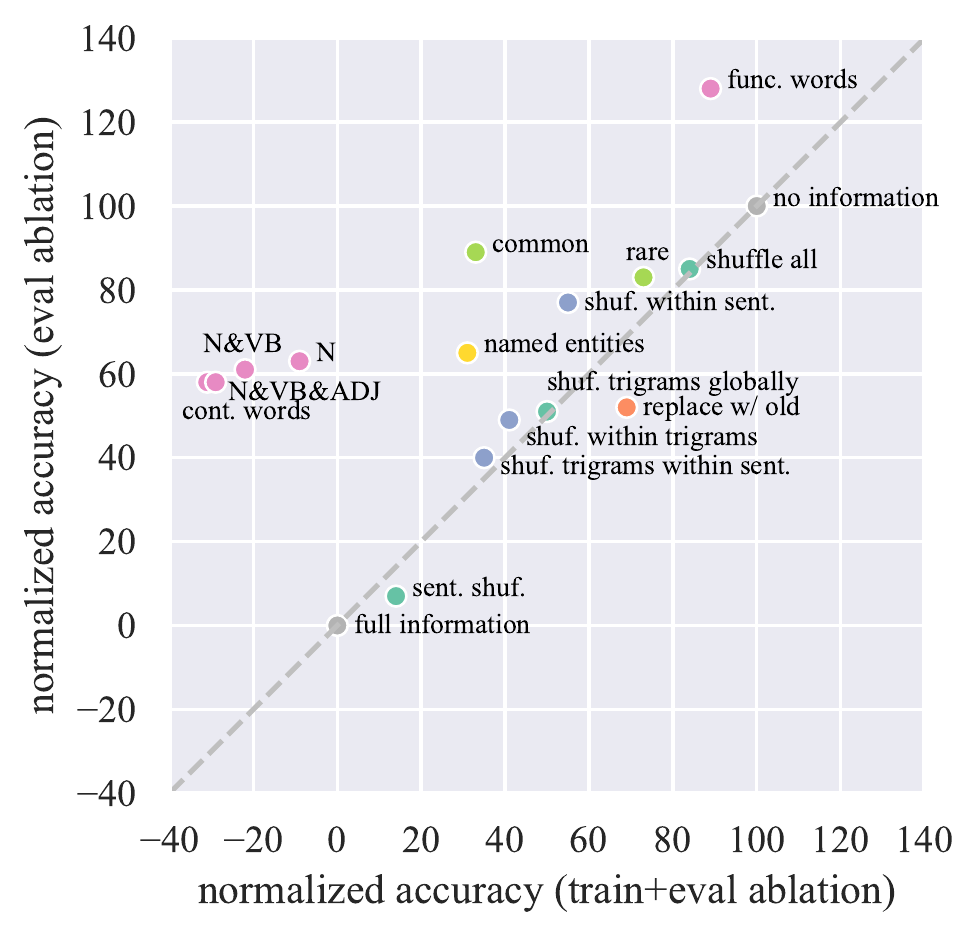} }}
    \caption{Comparison of model performance in the train+eval and eval-only settings. The units represent the percentage of the gap between the full information and no information models/contexts. That way, if a point falls on the dotted $y=x$ line, then that ablation has the same relative effect in each paradigm. If a point falls above the dotted line, then that ablation leads to better relative performance in the train+eval paradigm, and if a point falls below the dotted line, then that ablation leads to better relative performance in the eval-only paradigm.}
    \label{fig:compare}
    \vspace{-1em}
\end{figure}

In Figure~\ref{fig:compare} we show the contrast between the experimental paradigm of Sections~\ref{sec:experiments:order}--\ref{sec:experiments:words} and that of Section~\ref{sec:experiments:eval}. Especially for the experiments involving parts of speech, we see a significant difference in both the quantitative and qualitative results across the two paradigms.

\section{Longer Context Window}
\label{sec:longer-context}

Here we report the results of repeating the experiments of Sections 3.1 and 3.2 with ablated contexts of size 1024 tokens instead of 512 tokens in order to verify that the behavior we observed is not specific to the size of context window we chose.

\begin{figure}[ht!]
    \centering
    \subfloat[\centering Mid-range condition (first 256 tokens after ablation) ]{{\includegraphics[width=0.5\textwidth]{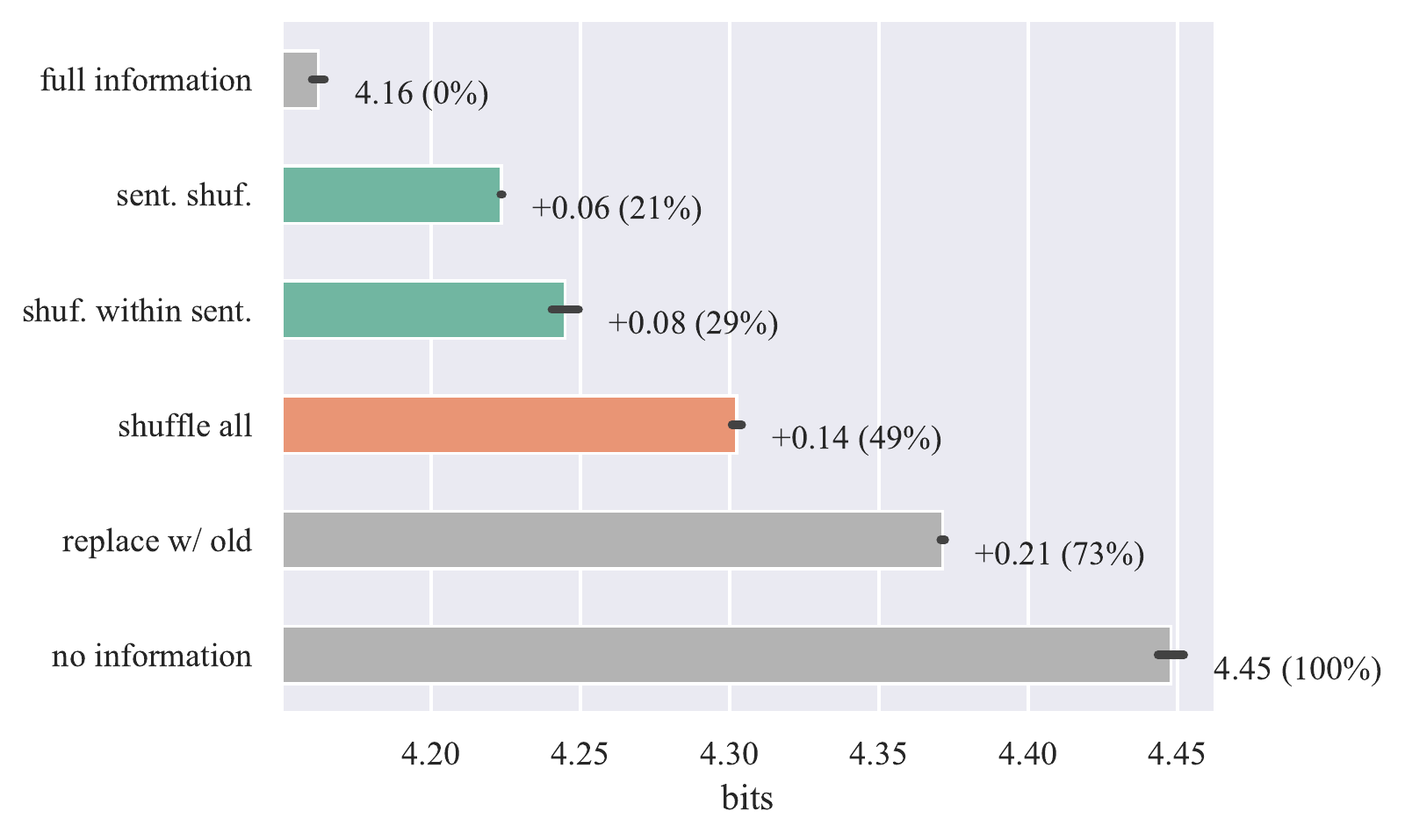} }}
    \\[-.5em]
    \hspace{-3mm}
    \subfloat[\centering Long-range condition (tokens 256-512 after ablation) ]{{\includegraphics[width=0.47\textwidth]{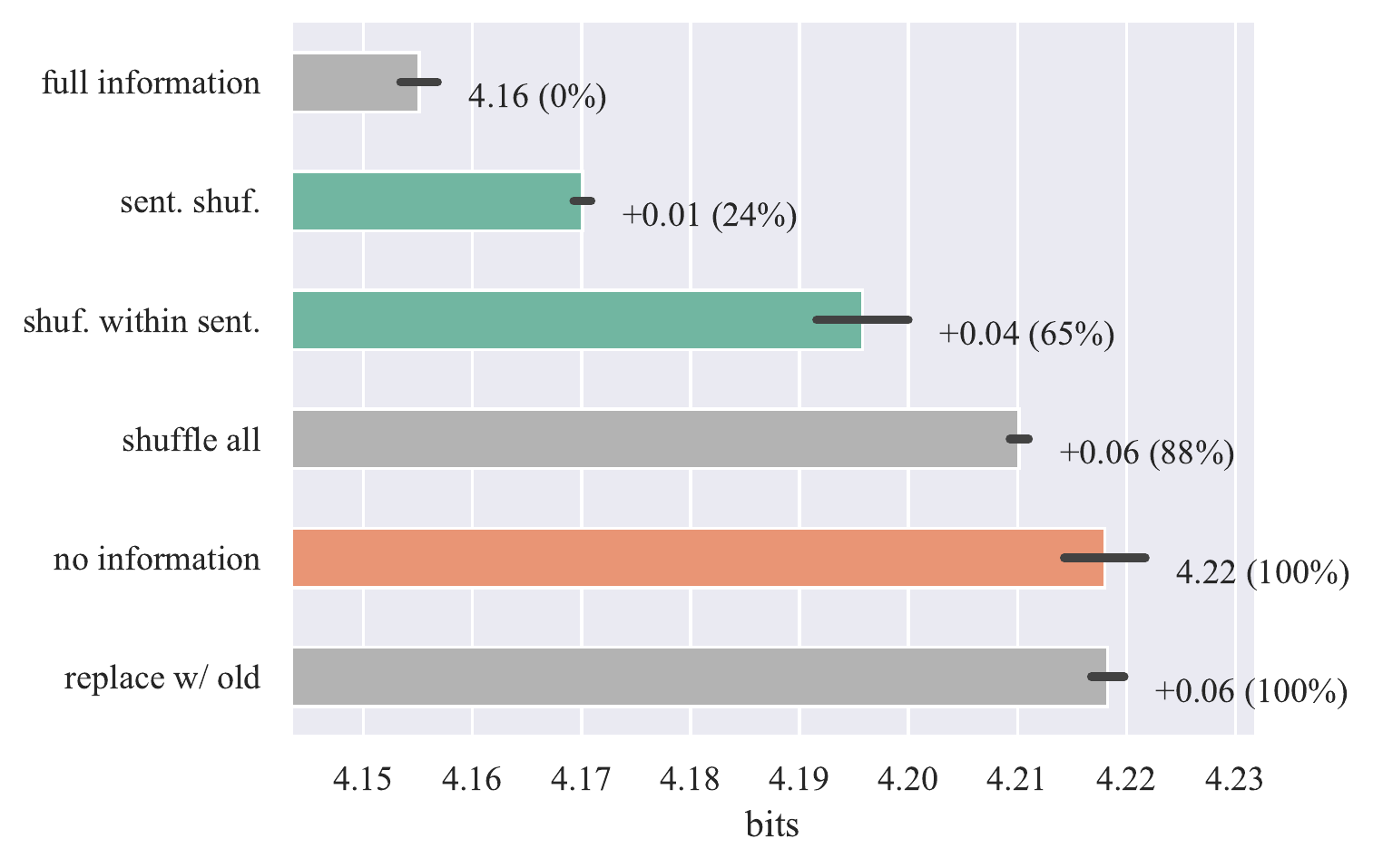} }}
    \caption{Effect of \textbf{word order} on usable information. Bar labels show ``change in ablated likelihood (ablated information)''. The $x$ axis shows ablated likelihood. Error bars represent 95\% confidence intervals. Ablated contexts contain 1024 tokens, but results are consistent with results on 512-token contexts.}
    \label{fig:1536-order}
    \vspace{-1em}
\end{figure}

\begin{figure}[ht!]
    \centering
    \subfloat[\centering Mid-range condition (first 256 tokens after ablation) ]{{\includegraphics[width=0.5\textwidth]{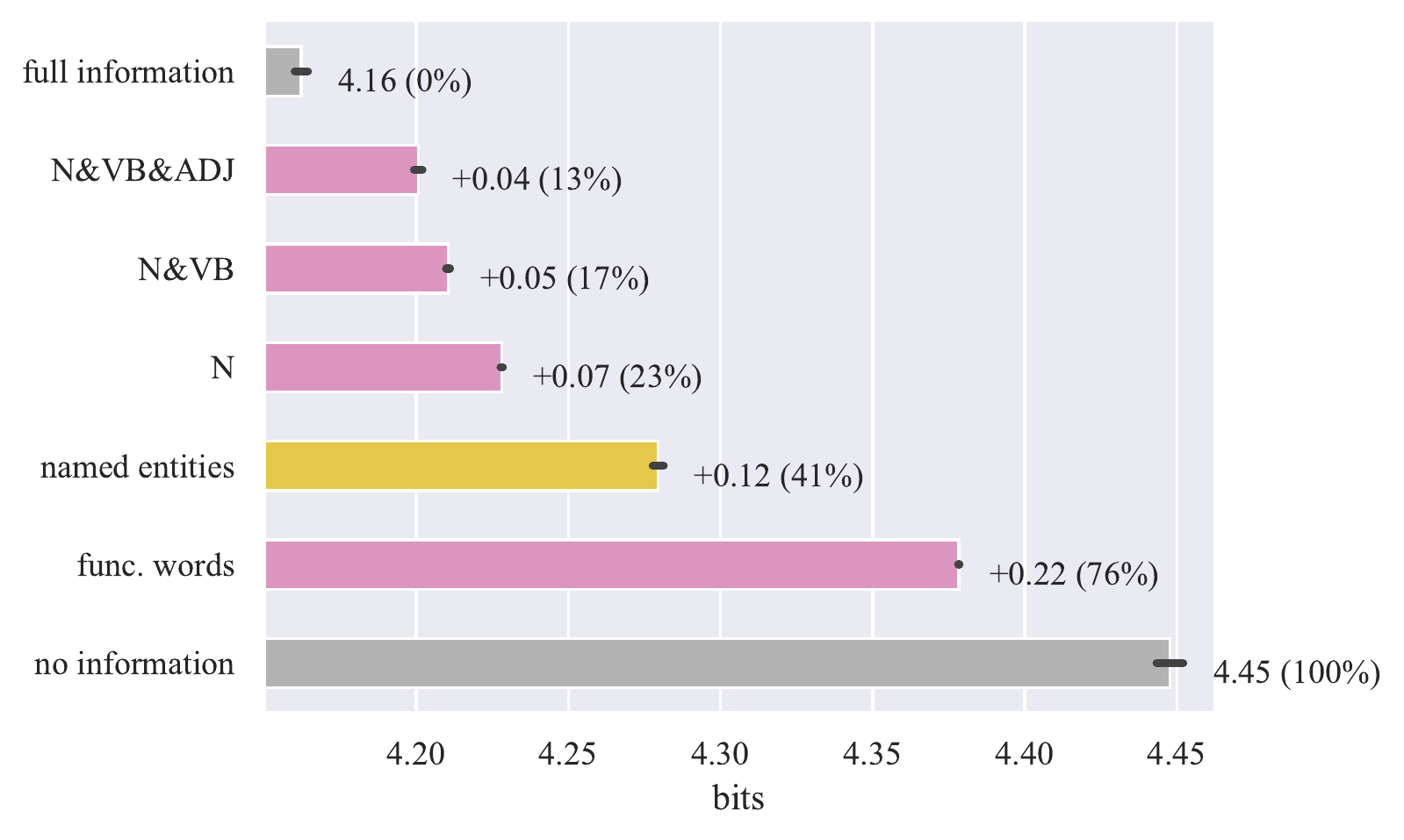} }}
    \\[-.5em]
    \hspace{-3mm}
    \subfloat[\centering Long-range condition (tokens 256-512 after ablation) ]{{\includegraphics[width=0.488\textwidth]{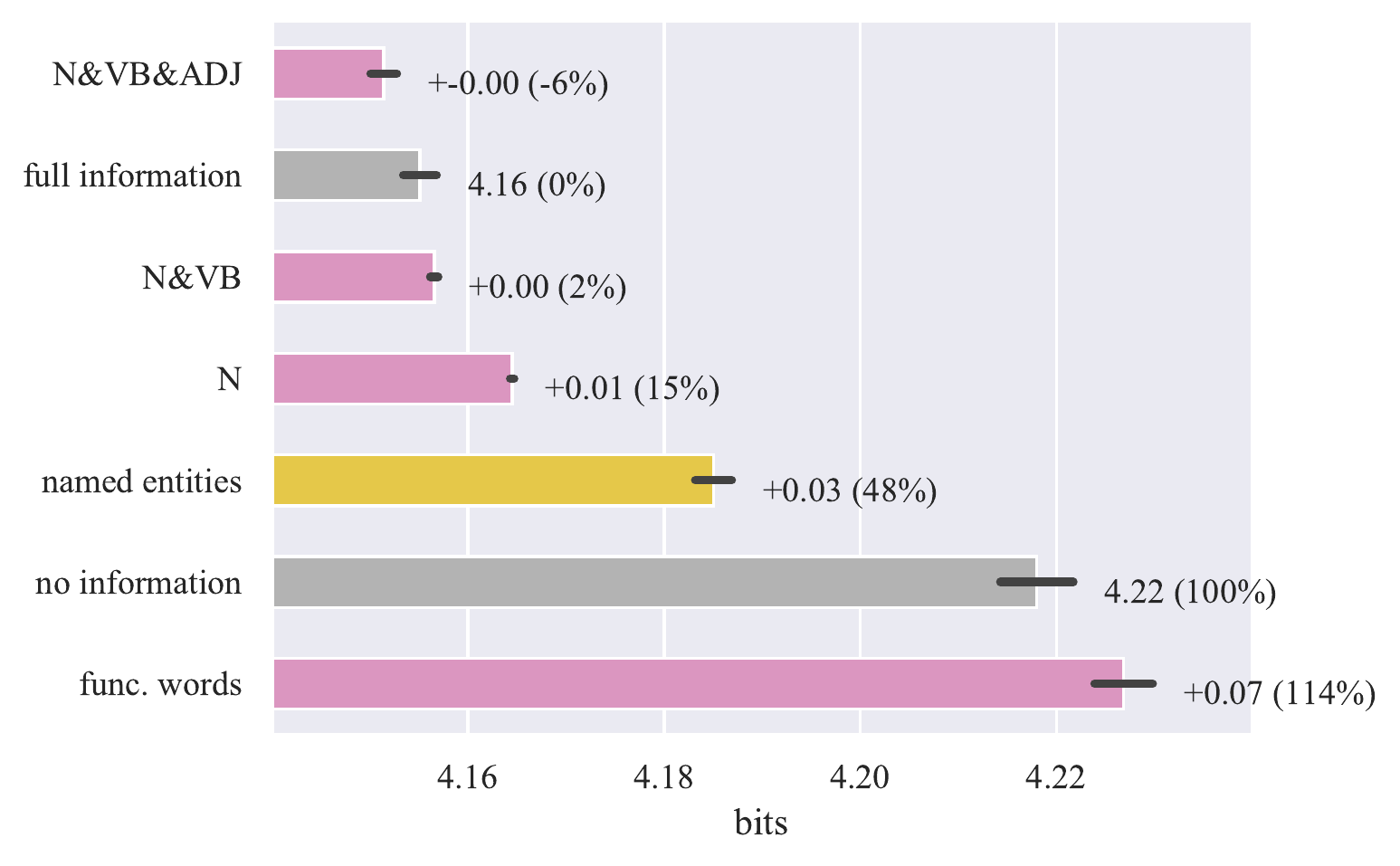} }}
    \caption{Effect of \textbf{word identity} on usable information. Labels are as in \cref{fig:1536-order}. Ablated contexts contain 1024 tokens, but results are consistent with results on 512-token contexts.}
    \label{fig:1536-word}
    \vspace{-1em}
\end{figure}

\section{Sample Generations}
\label{sec:generations}

The purpose of this section is to verify that models trained on ablated contexts can still generate text that is comparable to text generated by a model trained with full contextual information. We select a prompt from a randomly chosen Wikipedia article in the WikiText-103 validation set; each model generates a sentence (after finishing the sentence in progress) given the appropriately ablated version of the prompt. The prompt consists of 768 tokens, the last 256 of which remain unchanged for all versions of the prompt, so that the ablations are in the long range relative to the point of generation. The prompt and generations are as follows:
\vspace{4pt}
\newline
\textbf{prompt:}

\begin{quote}
at two independent schools for boys: Sussex House School, a day school in Chelsea's Cadogan Square, and the City of London School, a day school on the North Bank of the River Thames in London's financial district (known as the City of London). Attending school became difficult for Radcliffe after the release of the first Harry Potter film, with some fellow pupils becoming hostile, though he says it was people just trying to "have a crack at the kid that plays Harry Potter" rather than jealousy.

As his acting career began to consume his schedule, Radcliffe continued his education through on-set tutors. He admitted he was not very good at school, considering it useless and finding the work "really difficult." He achieved A grades in the three AS-level exams that he took in 2006, but decided to take a break from education and did not go to college or university. Part of his reasoning was that he already knew he wanted to act and write, and that it would be difficult to have a normal college experience. "The paparazzi, they'd love it," he told Details magazine in 2007. "If there were any parties going on, they'd be tipped off as to where they were."

= = Career = =

= = = Harry Potter = = =

In 2000, producer David Heyman asked Radcliffe to audition for the role of Harry Potter for the film adaptation of Harry Potter and the Philosopher's Stone, the best-selling book by British author J.K. Rowling. Rowling had been searching for an unknown British actor to personify the character, and the movie's director Chris Columbus recalled thinking, "This is what I want. This is Harry Potter", after he saw a video of the young actor in David Copperfield. Eight months later, and after several auditions, Radcliffe was selected to play the part. Rowling also endorsed the selection saying, "I don't think Chris Columbus could have found a better Harry." Radcliffe's parents originally turned down the offer, as they had been told that it would involve six films shot in Los Angeles. Warner Bros. instead offered Radcliffe a two-movie contract with shooting in the UK; Radcliffe was unsure at the time if he would do any more than that.

The release of Harry Potter and the Philosopher's Stone (released as Harry Potter and the Sorcerer's Stone in the United States) took place in 2001. Radcliffe received a seven figure salary for the lead role, but asserted that the fee was "not that important" to him; his parents chose to invest the money for him. The film was highly popular and was met with positive reviews, and critics took notice of Radcliffe: "Radcliffe is the embodiment of every reader's imagination. It is wonderful to see a young hero who is so scholarly looking and filled with curiosity and who connects with very real emotions, from solemn intelligence and the delight of discovery to deep family longing, "wrote Bob Graham of the San Francisco Chronicle.

A year later Radcliffe starred in Harry Potter and the Chamber of Secrets, the second installment of the series. Reviewers were positive about the lead actors' performances but had polarised opinions on the movie as a whole. The 2004 release Harry Potter and the Prisoner of Azkaban was the third film in the series. Radcliffe 's performance was panned by New York Times journalist A.O. Scott, who wrote that Watson had to carry him with her performance. Next was Harry Potter and the Goblet of Fire in 2005. The film was the second-highest gross
\end{quote}

\textbf{full information:} \begin{quote}
sing Potter film in the series at the time of its release.
After appearing in Harry Potter and the Philosopher's Stone, Radcliffe tried to avoid trying to improve his role.
\end{quote}
 
\textbf{shuffle all:} \begin{quote}
sing Potter film in the series (behind Harry Potter and the Chamber of Secrets), the highest grossing Harry Potter film in the series. The following year, Harry Potter and the Goblet of Fire was the top-grossing Potter film, making it the highest grossing Harry Potter film in the series.
\end{quote}
 
\textbf{shuf. within sent.:} \begin{quote}
sing Potter film in the series (until 2009's Gold Glove), selling 1.5 million tickets in its first week. The title of the film's sequel was also a major factor in the success of Philosopher, while its underperformance dropped because of series 3's Sci-Fi Channel.
\end{quote}

\textbf{shuf. within trigrams:} \begin{quote}
sing Potter film at the time of its release. The film had a financial success worldwide and received five Academy Award nominations at the 56th Academy Awards.
\end{quote}

\textbf{shuf. trigrams within sent.:} \begin{quote}
sing Potter film in the series at the United States box office. His appearance in Harry Potter and the Philosopher's Stone did not make it to the United States until 2006.
\end{quote}

\pagebreak

\textbf{shuf. trigrams globally:} \begin{quote}
sing Potter film at the time and was the third highest grossing film of 2006, the highest-grossing film in the series. In 2006, Radcliffe acted in his first major role in Harry Potter and the Deathly Hallows – Part I, where he played Harry's Chief Engineer, Harry Potter's primary antagonist.
\end{quote}
 
\textbf{sent. shuf.:} \begin{quote}
sing Potter film in the series at the time of its release. At the end of 2009, new backers were looking for a leading role in the film, with Radcliffe starting the role of Harry Potter and the Philosopher's Stone.
\end{quote}

\textbf{cont. words:} \begin{quote}
sing Harry Potter film of all time after Harry Potter and the Goblet of Fire, the highest-grossing film of the 2007–2008 financial year. The film's success meant that he and his wife bought a four-bedroom apartment in the 1960s, ``ready to keep them warm'', adding ``They could never make another effort to make something exciting for people who were usually very aware of the housing shortage but hadn't been very conscious of that''.
\end{quote}
 
\textbf{N \& VB \& ADJ:} \begin{quote}
sing animated film in the series after The Lion King. A similar amount of praise came in 2008 when The Magic King was released.
\end{quote}
 
\textbf{N \& VB:} \begin{quote}
sing animated film in the series. A significant amount of critical reviews were given by publications in The Washington Post, San Francisco Chronicle, and The Washington Post. The film was released in theaters in 2008.
\end{quote}
 
\textbf{N:}\begin{quote}
sing animated film in the series.

= Appearances = =
\end{quote}

\pagebreak

\textbf{named entities:} \begin{quote}
sing animated film in the series.

= = Appearances extended Persons (1990 – 2014) = =
\end{quote}

\textbf{rare:} \begin{quote}
sing animated film in the series.

= = Part two = =
\end{quote}

\textbf{common:} \begin{quote}
sing animated film in the series. A review in The New York Times found that Hans was not as strong as Watson but as well as Mr. Trough and Mr. Trough.
\end{quote}

\end{document}